\documentclass{svproc}
\usepackage{amsmath,amsfonts}
\usepackage{algorithmic}
\usepackage{algorithm}
\usepackage{hyperref}
\usepackage{array,multirow}
\usepackage[caption=false,font=normalsize,labelfont=sf,textfont=sf]{subfig}
\usepackage{textcomp}
\usepackage{siunitx}
\usepackage{url}
\usepackage{booktabs}
\usepackage{verbatim}
\usepackage{graphicx}
\usepackage{stfloats}
\usepackage[misc]{ifsym}

\newcommand{\compressParag}{\looseness=-1}

\begin{document}
\mainmatter              
\title{Model Predictive Contouring Control for Vehicle Obstacle Avoidance at the Limit of Handling}
\titlerunning{MPCC for Obstacle Avoidance}  
%
\author{Alberto Bertipaglia\inst{1} \textsuperscript{(\Letter)} \textsuperscript{[0000-0003-0364-8833]} \and Mohsen Alirezaei\inst{2} \textsuperscript{[0000-0002-2858-6408]}
Riender Happee\inst{1} \textsuperscript{[0000-0003-4530-8853]} \and Barys Shyrokau\inst{1} \textsuperscript{[0000-0003-4530-8853]}}
\authorrunning{Alberto Bertipaglia et al.} 
%
\tocauthor{Alberto Bertipaglia, Mohsen Alirezaei, Riender Happee, Barys Shyrokau}
\institute{Delft University of Technology, Delft, The Netherlands\\
\email{A.Bertipaglia@tudelft.nl}
\and 
Siemens PLM Software, Helmond, The Netherlands}

\maketitle              

\begin{abstract}
This paper proposes a non-linear Model Predictive Contouring Control (MPCC) for obstacle avoidance in automated vehicles driven at the limit of handling. The proposed controller integrates motion planning, path tracking and vehicle stability objectives, prioritising obstacle avoidance in emergencies. The controller's prediction model is a non-linear single-track vehicle model with the Fiala tyre to capture the vehicle's non-linear behaviour. The MPCC computes the optimal steering angle and brake torques to minimise tracking error in safe situations and maximise the vehicle-to-obstacle distance in emergencies. Furthermore, the MPCC is extended with the tyre friction circle to fully exploit the vehicle's manoeuvrability and stability. The MPCC controller is tested using real-time rapid prototyping hardware to prove its real-time capability. The performance is compared with a state-of-the-art Model Predictive Control (MPC) in a high-fidelity simulation environment. The double lane change scenario results demonstrate a significant improvement in successfully avoiding obstacles and maintaining vehicle stability. \compressParag
\end{abstract}
\section{Introduction}
Automated vehicles' safety relies heavily on their ability to effectively avoid obstacles through evasive manoeuvres. Nevertheless, tyre non-linearities pose a significant challenge in this regard. \cite{brown2019coordinating}. A hierarchical controller architecture typically separates motion planning, path tracking, and vehicle stability tasks \cite{Falcone2008Hiera}. Although each task can be optimised separately, in an evasive manoeuvre, the three objectives could be in conflict with each other \cite{funke2016collision}. Therefore, to avoid potential conflict,  we integrate motion planning, path tracking, and vehicle stability into a single controller for obstacle avoidance at high speed.\compressParag\newline \indent
Recent studies highlight that vehicle stability constraints could lead to tracking errors, potentially causing collisions \cite{funke2016collision,lenssen2023combined}. A potential solution is integrating an obstacle avoidance controller with objectives including motion planning, path tracking and vehicle stability \cite{Gao2014Robust}. Model Predictive Control (MPC) based on a non-linear single-track vehicle model can integrate all the controllers' objectives and modify the desired trajectory to keep the vehicle stable and at a safe distance from the object \cite{funke2016collision}. In order to run the MPC controller in real-time, the non-linear single-track vehicle model is linearised into an affine time-varying model. The longitudinal and lateral control is considered separately. However, the affine model diminishes the model's accuracy and limits the control capabilities. Furthermore, the model fidelity is particularly affected when the lateral and longitudinal dynamics are coupled \cite{brown2019coordinating}. Thus, to address these limitations, a non-linear vehicle model and non-linear optimisation must be adopted \cite{brown2019coordinating,chowdhri2021integrated}. The vehicle and the obstacles are represented as a set of circles, and their distance is constantly measured in the cost function of the non-linear MPC \cite{brown2019coordinating}. Vehicle kinematics is described using the Frenet coordinate system because it allows an easy determination of the vehicle location relative to a reference line. Despite these advantages, when the trajectory has a curvature, the vehicle-to-obstacle (V2O) distance in the Frenet coordinate system is over-estimated with respect to the distance in the Cartesian coordinate system. Furthermore, the computation of the travelled distance of the vehicle with respect to the reference line at every time step requires an additional optimisation \cite{liniger2015optimization}.\newline\indent
This paper proposes a Model Predictive Contouring Control (MPCC) based on a non-linear single-track vehicle model for obstacle avoidance. The MPCC, recently proposed for robot motion planning at low speed \cite{brito2019model} and lap-time optimisation for scaled vehicle \cite{liniger2015optimization}, is extended to consider obstacle avoidance and vehicle stability. The MPCC, using a Cartesian frame, aims to approximate the MPC based on the Frenet reference system, thanks to the introduction of the lag and contouring error in the cost function. Thus, it avoids overestimating the V2O distance due to the Frenet coordinates, and it eliminates the additional optimisation to compute the travelled distance of the vehicle with respect to the reference line. We exhaustively assess the controller performance in a high-fidelity simulation environment designing a double lane change for vehicle obstacle avoidance.\newline\indent
The main contribution of this paper is twofold. The first is improving the overall obstacle avoidance performance of the proposed MPCC over a state-of-the-art MPC \cite{brown2019coordinating}. Both controllers can successfully avoid a collision between the vehicle and the obstacle in a double-lane change. However, the baseline MPC cannot keep the vehicle outside the unsafe area close to obstacles or road edges. The second contribution encompasses improved vehicle stability by minimising the peaks in sideslip angle and increasing the minimum velocity during manoeuvres due to better prioritisation of the obstacle avoidance objective within the MPCC framework.\compressParag

\section{Prediction Model}
A non-linear single-track vehicle model is used in the proposed MPCC. Only the in-plane dynamics are considered, ignoring the lateral weight transfer and the roll and pitch dynamics. The vehicle position is described using a Cartesian reference frame by the states $\left(x = [X, Y, \phi, v_x, v_y, r, \theta, \delta, F_x]\right)$: longitudinal position $\left(X \right)$, lateral position $\left(Y \right)$, and the heading angle $\left(\phi \right)$ of the vehicle centre of gravity (CoG) relative to an inertial frame. The velocity states are the longitudinal and lateral velocity at the CoG, respectively $\left(v_x\right)$ and $\left(v_y\right)$, and the yaw rate $\left(r\right)$. Furthermore, an additional state corresponding to the vehicle travelled distance $\left(\theta\right)$ is introduced, and the MPCC cost function uses it to compute the vehicle position relative to the reference line. The steering angle $\left(\delta\right)$ and the longitudinal force $\left(F_x\right)$ correspond to the integral of the control inputs. The implemented state derivatives, $\left(\dot{x}\right)$, correspond to the following equations:
\begin{equation}
    \begin{split}
    \begin{cases}
        \dot{X} = v_x \cos \left(\psi\right) + v_y \sin \left(\psi\right)\\
        \dot{Y} = v_x \sin \left(\psi\right) + v_y \cos \left(\psi\right)\\
        \dot{\psi} = r\\
        \dot{v}_x = \frac{- F_{yf}\left(x, u_v\right) \sin \left(\delta \right) + F_{xf}\left(x, u_v\right) \cos \left(\delta\right) + F_{xr}\left(x, u_v\right) - F_{drag}}{m} + r v_y\\
        \dot{v}_y = \frac{F_{yf}\left(x, u_v\right) \cos \left(\delta \right) + F_{xf} \sin \left(\delta\right) + F_{yr}\left(x, u_v\right)}{m} - r v_x\\
        \dot{r} = \frac{l_f F_{yf}\left(x, u_v\right) \cos \left(\delta \right) +  l_f F_{xf}\left(x, u_v\right) \sin \left(\delta\right) - l_r F_{yr}\left(x, u_v\right) }{I_{zz}}\\
        \dot{\theta} = \sqrt{v_x^2 + v_y^2}
    \end{cases}
    \end{split}
    \label{eq:Single}
\end{equation}
where $F_{xi}$ and $F_{yi}$  are, respectively, the longitudinal and lateral tyre forces, $i$ stands for front $\left(f\right)$ or rear $\left(r\right)$, $l_f$ and $l_r$ are the distance between the front and rear axle to the CoG, $I_{zz}$ is the vehicle inertia around the z-axis, $m$ corresponds to the vehicle mass and $F_{drag}$ is the aerodynamic drag resistance.\newline\indent
The vehicle model inputs $\left(u_v\right)$ are the road-wheel angle rate $(\dot{\delta})$, the total longitudinal force rate applied at the CoG $(\dot{F}_x)$ and the brake repartition between the front and rear axle $\left(\lambda_b\right)$. The control input rates are integrated into the prediction model before being applied to the vehicle. The inputs governing the longitudinal dynamics are $F_x$ and $\lambda_b$ and not $F_{xf}$ and $F_{xr}$. The reason is that the vehicle must be able to accelerate and brake, but it is forbidden to accelerate and brake simultaneously \cite{brown2019coordinating}. Thus, the constraint $\left(F_{xf} F_{xr} \geq 0\right)$ is commonly applied. However, it introduces a saddle point when both $F_{xr}$ and $F_{xr}$ are equal to zero, not guaranteeing the Hessian to be positive-definite \cite{brown2019coordinating}. For this reason, the constraint is implicitly formulated inside the prediction model as follows:
\begin{equation}
    \begin{split}
        F_{xf} = 
        \begin{cases}
            \lambda_b F_x & \text{if } F_x \leq 0 \\
            \lambda_d F_x, & \text{otherwise}
        \end{cases} \;\;\;\;\;\;\;\;\;\;\;\;\;\;\;
        F_{xr} = 
        \begin{cases}
            \left( 1 - \lambda_b \right) F_x & \text{if } F_x \leq 0 \\
            \left( 1 - \lambda_d \right)  F_x, & \text{otherwise}
        \end{cases}
    \end{split}
    \label{eq:ifcases}
\end{equation}
where $\lambda_d$ represents whether the vehicle is front- or rear-wheel driven. \newline\indent
A Fiala tyre model computes the lateral tyre force for each axle, while the longitudinal force is defined as an input of the system \cite{brown2019coordinating}. The non-linear coupling between $F_{xi}$ and $F_{yi}$ is captured according to the friction circle \cite{chowdhri2021integrated}. The tyre parameters are optimised by performing quasi-steady-state circular driving in a high-fidelity simulation based on a Delft-Tyre model 6.1.\newline\indent
The vehicle and obstacles are represented as circles so that their Euclidean distance can constantly be computed as follows:
\begin{equation}
    D_{V2O} = \sqrt{\left( X - X_{obs}\right)^2 + \left( Y - Y_{obs}\right)^2} - r_{obs} - r_{veh}
    \label{eq:dista}
\end{equation}
where $X,\, Y$ and $X_{obs},\, Y_{obs}$ are, respectively, the longitudinal and lateral position of the vehicle and obstacle centre, and $r_{veh}$ and $r_{obs}$ are the radius of the vehicle and obstacle circles. Thus, the vehicle and the obstacle collide when the V2O distance $D_{V2O}$ is lower than zero. The controller aims to keep $D_{V2O}$ always positive and above a user-defined safety distance.
\section{Model Predictive Contouring Control}
This section describes how the proposed MPCC controller integrates motion planning, path tracking, and vehicle stability objectives. 
\subsection{Cost Function}
The MPCC employs an iterative approach to solve an optimal control problem, enabling the vehicle to operate at its handling limit while avoiding obstacles. The cost function objectives encompass several aspects: tracking a reference longitudinal and lateral position, maintaining a desired velocity, dynamically adjusting the indicated trajectory to ensure a safe distance from obstacles while maintaining stability, and guaranteeing the physical feasibility of input signals. The proposed cost function $\left(J\right)$ is:
\begin{equation}
    \begin{split}
        J = \sum_{i=1}^{N} \biggl( & q_{e_{Con}} e_{Con, i}^2 + q_{e_{Lag}} e_{Lag, i}^2 + q_{e_{Vel}} \left(e_{Vel}\right)^2 + q_{\dot{\delta}} \dot{\delta_i}^2 + q_{\dot{F_x}} \dot{F_{x_i}}^2 +\\
        & + q_{\lambda_b} e_{\lambda_{b, i}}^2 + \sum_{j=1}^{N_{obs}}\left( q_{e_{V2O}} e_{V2O, j, i}^2 \right) + \sum_{j=1}^{N_{edg}}\left( q_{e_{V2E}} e_{V2E, j, i}^2 \right) \biggr) 
    \end{split}
    \label{eq:cost}
\end{equation}
where $N$ is the length of the prediction horizon, $N_{obs}$ is the number of obstacles in the road, $N_{edg}=2$ is the number of road edges, and parameters $q_*$ are the weights of the respective quadratic errors, defined below.  The weights are fine-tuned to optimise controller performance by minimising longitudinal velocity error and sideslip angle peaks \cite{Bertipaglia2022Two} and enabling the vehicle to avoid obstacles maintaining a safe distance from obstacles and road edges.\newline\indent
\begin{figure}[t]
    \centering
    \includegraphics[height = 2.9 cm,keepaspectratio]{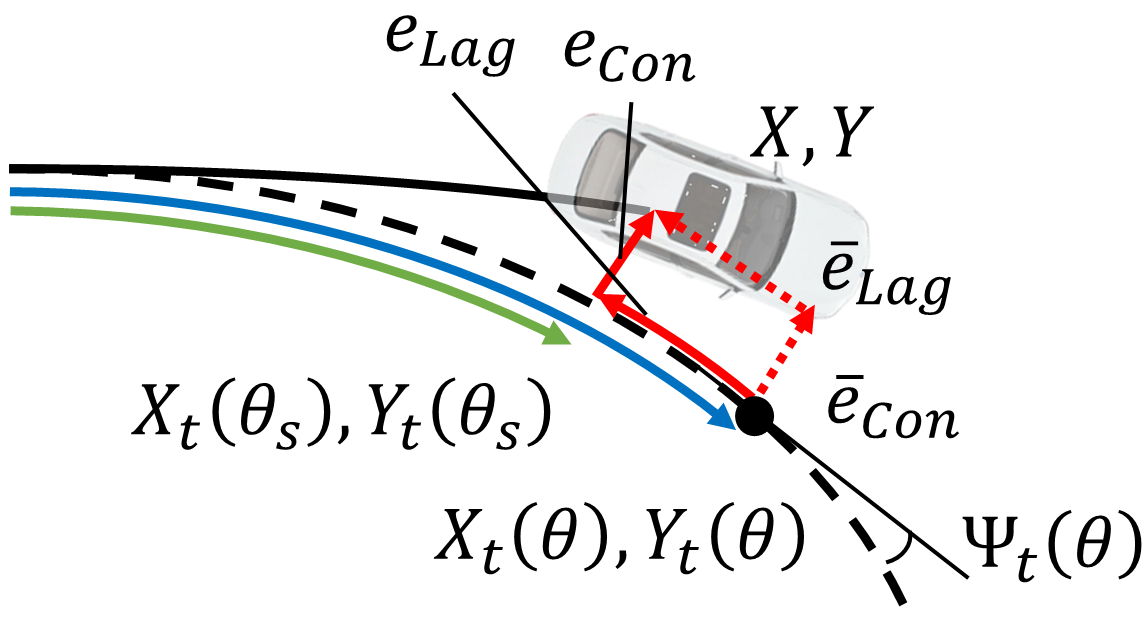}
    \caption{A representation of the contouring $e_{Con}$ and lag error $e_{Lag}$.
    $\theta$ and $\theta_s$ are the vehicle travelled distance and the distance with respect to the reference line.}
    \label{fig:eCeL}
\end{figure}
The reference trajectory is tracked through the introduction of the contouring error $\left(e_{Con}\right)$ and the lag error $\left(e_{Lag}\right)$ \cite{liniger2015optimization,brito2019model}, see Fig. \ref{fig:eCeL}. $e_{Con}$ corresponds to the projection of the vehicle position over the desired trajectory. It is computed as a function of the vehicle travelled distance with respect to the reference line $\left(\theta_s\right)$. However, based on the Cartesian coordinate frame, the controller cannot determine the distance $\theta_s$ in the prediction model of eq. \ref{eq:Single}. Vice versa, it corresponds to a state when a Frenet reference system is employed \cite{brown2019coordinating,chowdhri2021integrated}. Thus in the MPCC, $\theta_s$ is approximated by the total vehicle travelled distance $\theta$ computed as in eq. \ref{eq:Single}. To ensure the validity of this approximation, the norm between the two distances, called lag error, must be minimised. The errors are approximated as follows \cite{liniger2015optimization,brito2019model}:\compressParag
\begin{equation}
    \begin{split}
        &\bar{e}_{Con} = \sin \left( \Psi_t\left(\theta\right)\right) \left(X - X_t\left( \theta \right) \right) - \cos \left( \Psi_t\left(\theta\right)\right) \left(Y - Y_t\left( \theta \right) \right)\\
        &\bar{e}_{Lag} = - \cos \left( \Psi_t\left(\theta\right)\right) \left(X - X_t\left( \theta \right) \right) - \sin \left( \Psi_t\left(\theta\right)\right) \left(Y - Y_t\left( \theta \right) \right)
    \end{split}
\end{equation}
where $X_t$, $Y_T$, and $\Psi_T$ are the desired longitudinal, lateral position and heading angle. It should be noted that $\bar{e}_{Lag}$ is minimised by following the reference trajectory and modifying the vehicle velocity. Thus, if the desired velocity is unfeasible for the given trajectory, the controller will reduce it to keep the $\bar{e}_{Lag}$ close to zero.\compressParag\newline\indent
The velocity is tracked by computing $e_{Vel}$, which corresponds to the difference between the vehicle's velocity and the desired one.\newline \indent
The controller dynamically adjusts the reference trajectory to ensure a safe distance from obstacles. This is possible by evaluating the V2O distance. The error function $\left(e_{V20} = D_{V2O} - D_{Sft, O}\right)$ is computed as the difference between a user-defined safety distance between the vehicle and the obstacle $\left(D_{Sft, O}\right)$, and the V2O distance $D_{V2O}$, see eq. \ref{eq:dista}. It should be noted that when $D_{V2O} \geq D_{Sft, O}$, $q_{V2O}$ is automatically set at zero, while when $D_{V2O} \leq D_{Sft, O}$ the error is computed to keep the vehicle always at a safe distance from the object. A similar error is introduced to keep the vehicle away from the road edges $\left(e_{V2E} = D_{V2E} - D_{Sft, E}\right)$. $D_{Sft, E}$ is the safety distance between the vehicle and the road edge, and $D_{V2E}$ is the distance between the vehicle and the road edge.\compressParag\newline\indent
The other cost terms focus on the smoothness and feasibility of control inputs. Thus, the cost terms are introduced to the steering angle rate and the longitudinal force rate applied at the vehicle CoG. Furthermore, the controller minimises the error $\left(e_{\lambda_b}\right)$ between the actual brake repartition and the ideal one, similar to the ideal brake force distribution. The ideal brake repartition, $\frac{F_{zf}}{F_{zr}}\slash\frac{F_{zf}+F_{zr}}{F_{zr}}$, corresponds to the percentage of the total braking force that should be applied to the front axle to lock the front and the rear axle simultaneously. The error $e_{\lambda_b}$ helps the controller provide the ideal braking repartition. It also allows a variation depending on various factors, e.g., the tyre saturation in the front and rear axle or the value of the steering angle.

\subsection{Constraints}
The cost function is constrained based on actuator limitations, vehicle stability and road track width. The road-wheel steering angle, the total longitudinal force and their respective rates are limited by upper and lower constraints. \newline\indent
The vehicle stability is enforced by restricting the total available tyre force at each axle using the tyre friction circle. At first, the longitudinal force $\left(F_x\right)$ is restricted considering $|F_x|\leq sf \mu F_z$. Given the difficulties of estimating the road friction coefficient $\left(\mu \right)$, the tyre friction circle is multiplied by a safety coefficient $\left(sf\right)$ equal to \SI{0.95}{}, which limits the available longitudinal force. The Fiala tyre in the prediction model limits the maximum lateral tyre force according to the tyre friction circle at a given $F_x$.\compressParag \newline\indent
The vehicle is enforced to stay on the track using the following constraint:
\begin{equation}
    \biggl\lVert 
    \begin{bmatrix}
        X\\
        Y
    \end{bmatrix}
    - 
    \begin{bmatrix}
        X_{cen}\\
        Y_{cen}
    \end{bmatrix}
    \biggr\rVert^2 \leq \left(\frac{W_{t}}{2}\right)^2 
\end{equation}
where $X_{cen}$ and $Y_{cen}$ are the longitudinal and lateral location of the centre of the track, and $W_{t}$ is the width of the road track \cite{liniger2015optimization}.

\subsection{Obstacle Avoidance Prioritisation}
The cost function combines different objectives and needs to prioritise collision avoidance during an evasive manoeuvre. Thus, the weights associated with the V2O or vehicle-to-edge (V2E) distance vary dynamically, according to:
\begin{equation}
    q_{V2O} = 
    \begin{cases}
        P_{k}, & \text{if } D_{V2O} < 0 \\
        P_{k}\; e^{-\frac{2 D_{V2O}^2}{D_{Sty, O}^2 }}, & \text{elseif}\;\; 0 \leq D_{V2O} \leq D_{Sty, O} \\
        0, & \text{otherwise}
    \end{cases} \\
\end{equation}
where $P_{k}$ represents the maximum value that $q_{V2O}$ can reach. This value prioritises collision avoidance as the V2O distance decreases. It increases up to $P_{k}$ in a Gaussian shape curve. This prioritisation approach is more gradual than the step growth from  \cite{brown2019coordinating}, which results in faster and more robust convergence of the MPCC solver. It is important to note that prioritising collision avoidance may result in a larger tracking error and deviation from the vehicle's desired velocity.
\section{Experimental Setup}
The proposed control is tested on the SCALEXIO dSPACE real-time platform, based on a multi-core DS6001 processor (\SI{2.8}{GHz} quad-core, \SI{1}{GB} DDR2 SD RAM). The MPCC is set up in a separate core from the vehicle plant. The prediction model is discretised using Runge-Kutta 2 for its proper trade-off between accuracy and simplicity \cite{brown2019coordinating}. A \SI{0.05}{s} sampling time and \SI{50}{} steps prediction horizon is selected to make the controller real-time implementable. The optimisation problem is solved using the non-linear interior point solver of FORCESPro \cite{FORCESPro}. The optimisation Hessian is approximated using the Broyden–Fletcher–Goldfarb–Shanno algorithm, and a user-defined initial Hessian matrix is provided to speed up the solving time. All the other solver parameters are kept as defaults. The platform successfully solves the non-linear optimisation with an average time of \SI{15.6}{ms} and max solving time of \SI{18.6}{ms}. However, there is no mathematical guarantee that the solver will converge in time. The vehicle plant runs in a separate core at \SI{1000}{Hz}. It is a high-fidelity BMW Series 545i vehicle model based on an IPG CarMaker simulation platform. Its parameters are determined through experimental inertia measurements, while the suspensions are characterised using a Kinematics \& Compliance test rig. The tyre dynamics are based on a Delft-Tyre 6.1. The actuator dynamics are included through a second-order transfer function to increase the simulation accuracy \cite{chowdhri2021integrated}.\compressParag\newline\indent
A double lane change manoeuvre with two obstacles is considered to assess the proposed controller's capabilities in avoiding obstacles at the limit of handling. The manoeuvre contains multiple obstacles because it is crucial to assess how the re-planning to avoid the first obstacle influences the vehicle’s capacity to avoid the second one. A coarse trajectory is provided, corresponding to the desired path from the behaviour planner. The desired path indicates which side of the obstacle the collision avoidance controller needs to follow. Thus, the proposed controller follows the desired trajectory while also making necessary adjustments when the trajectory is too close to an obstacle or deemed infeasible. Considering the initial vehicle position and orientation, the desired trajectory is determined as a function of the road curvature and the distance along the reference line. The desired vehicle velocity is constant along all the manoeuvres \cite{brown2019coordinating}.
\section{Results}
This section compares and analyses the performance of the MPCC with a baseline MPC for collision avoidance \cite{brown2019coordinating}.
\subsection{Cartesian vs Frenet Reference Frame}
\begin{figure}[t]
    \centering
    \captionsetup[subfloat]{font=tiny}
    \subfloat[Scenario in Cartesian coordinates.\label{fig:CirclCart}]{%
       \includegraphics[width=0.25\textwidth, keepaspectratio]{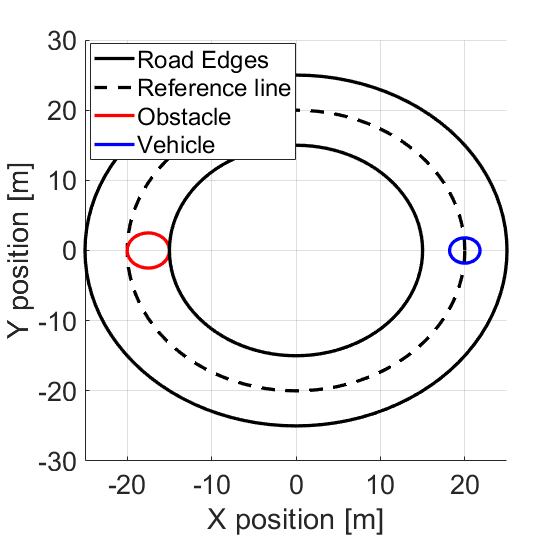}} 
    \subfloat[Scenario in Frenet coordinates.\label{fig:CirclCurv}]{%
       \includegraphics[width=0.25\textwidth, keepaspectratio]{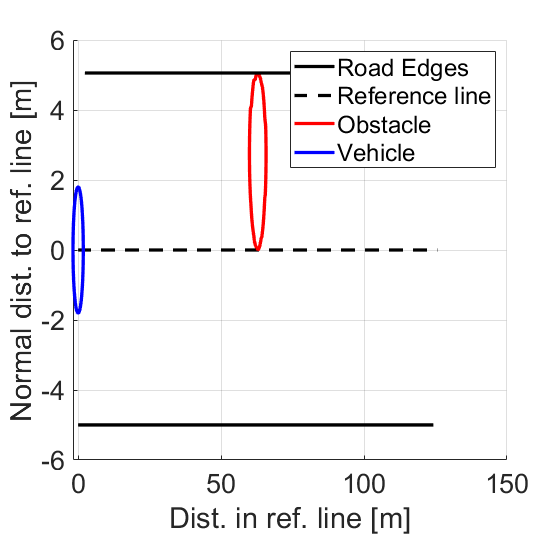}}
    \subfloat[\SI{20}{m} radius trajectory.\label{fig:Circle20}]{%
       \includegraphics[width=0.25\textwidth, keepaspectratio]{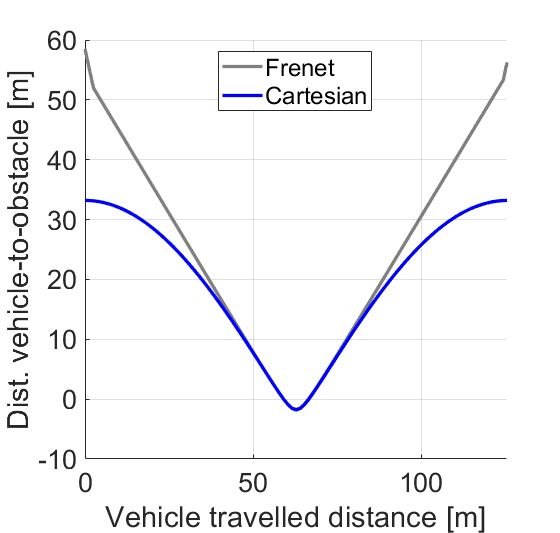}} 
    \subfloat[V20 overestimation in Frenet frame.\label{fig:CirclesDista}]{%
       \includegraphics[width=0.25\textwidth, keepaspectratio]{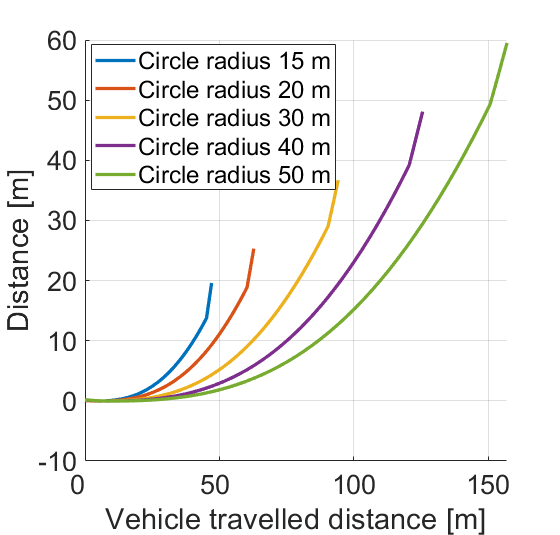}} \hfill
    \caption{Fig. \ref{fig:CirclCart} and Fig. \ref{fig:CirclCurv} show the scenario in Cartesian and Frenet coordinates. Fig. \ref{fig:Circle20} - the V2O distance in a \SI{20}{m} radius circular trajectory using Frenet coordinates. Fig. \ref{fig:CirclesDista} - effect of trajectory curvature on the V2O distance overestimation.\compressParag}
    \label{fig:Circle}
\end{figure}

The proposed MPCC controller computes the V2O and V2E distances using a Cartesian reference system. Vice versa, the baseline is an MPC built on a Frenet reference frame. The difference in the coordinate system implies a disagreement on how distance is measured. Fig. \ref{fig:CirclCart} and Fig. \ref{fig:CirclCurv} show a vehicle driven on a \SI{20}{m} radius circular road in the Cartesian and Frenet reference system. An obstacle is located on one side of the circle, with a normal displacement from the reference line equal to the radius of the obstacle. Assuming the vehicle drives on the reference line, the V2O distance is computed using both coordinates at every instance. Fig. \ref{fig:Circle20} shows how the two distances vary depending on the vehicle's distance driven on the reference line. The distance in Frenet coordinate frame \cite{brown2019coordinating} is similar to the one computed in the Cartesian coordinate frame only when the vehicle is close to the obstacle. Vice versa, when the vehicle still needs to drive along the reference line, the V2O distance in the Frenet reference frame is overestimated compared with the Cartesian distance. Thus, the MPCC can prioritise the obstacle avoidance objective before the MPC based on the Frenet reference frame, improving the chances of safely avoiding the obstacle while keeping the vehicle stable. The difference between the Frenet and the Cartesian V2O distances depends on the road's curvature and the obstacle's normal distance from the reference line. Fig. \ref{fig:CirclesDista} shows how the difference in V2O distance between Frenet and Cartesian changes with the road curvature. It can be observed that the higher the road curvature, the greater will be the overestimation.

\subsection{Double Lane Change}
\begin{figure}[t]
    \centering
    \captionsetup[subfloat]{font=tiny}
    \subfloat[Vehicle trajectories. \label{fig:DLTraje}]{%
       \includegraphics[width=1\textwidth, keepaspectratio]{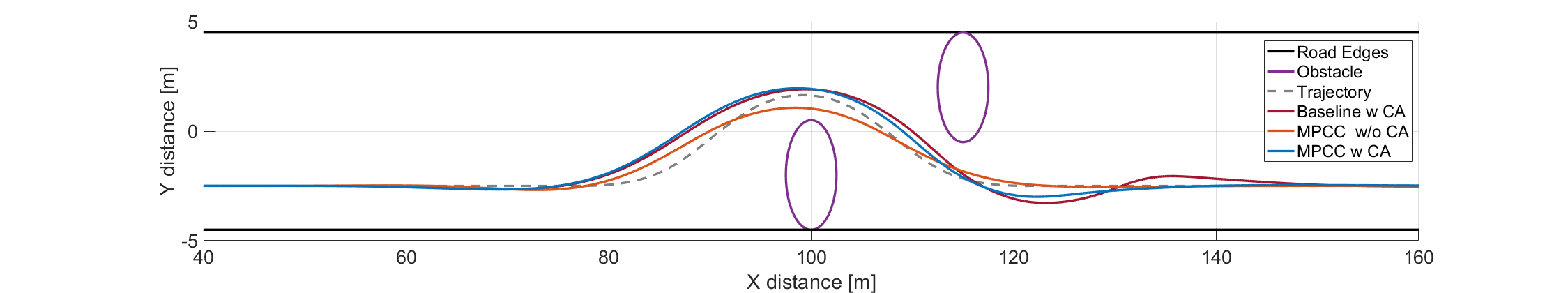}} \hfill \\
    \subfloat[Vehicle-to-edge distance.\label{fig:DLRE}]{%
       \includegraphics[width=0.32\textwidth, keepaspectratio]{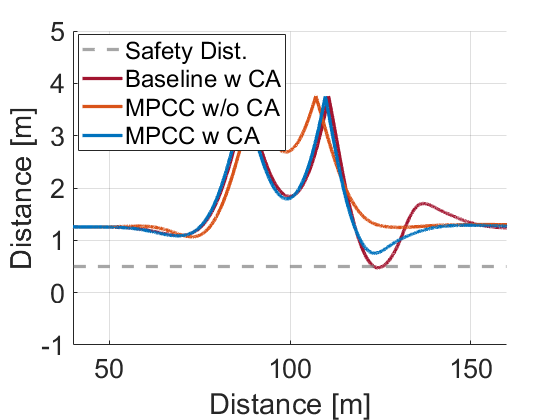}} 
    \subfloat[Vehicle-to-obstacle distance.\label{fig:DLRO}]{%
       \includegraphics[width=0.32\textwidth, keepaspectratio]{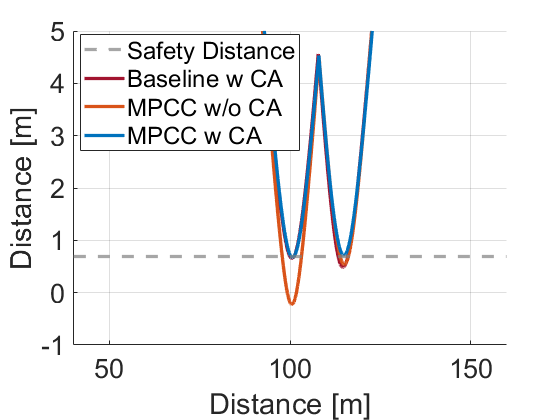}} 
    \subfloat[Longitudinal velocity.\label{fig:DLVel}]{%
       \includegraphics[width=0.32\textwidth, keepaspectratio]{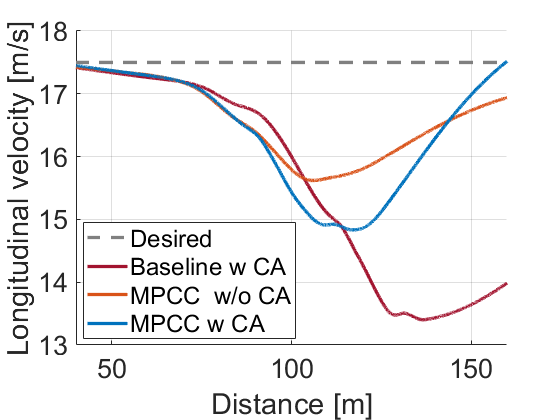}} \hfill \\
    \subfloat[Vehicle sideslip angle.\label{fig:DLBeta}]{%
       \includegraphics[width=0.25\textwidth, keepaspectratio]{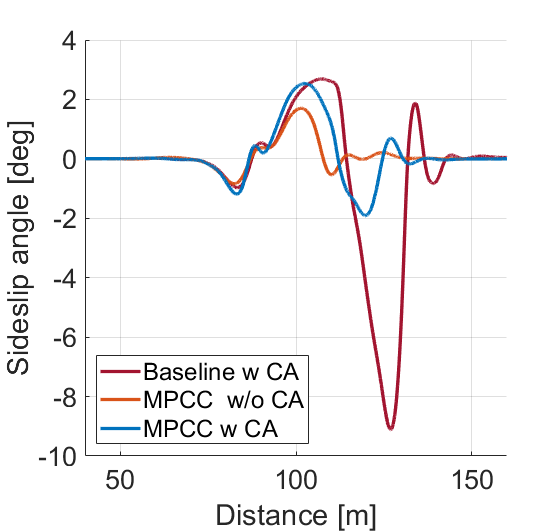}}
    \subfloat[Road-wheel angle.\label{fig:DLDelta}]{%
       \includegraphics[width=0.25\textwidth, keepaspectratio]{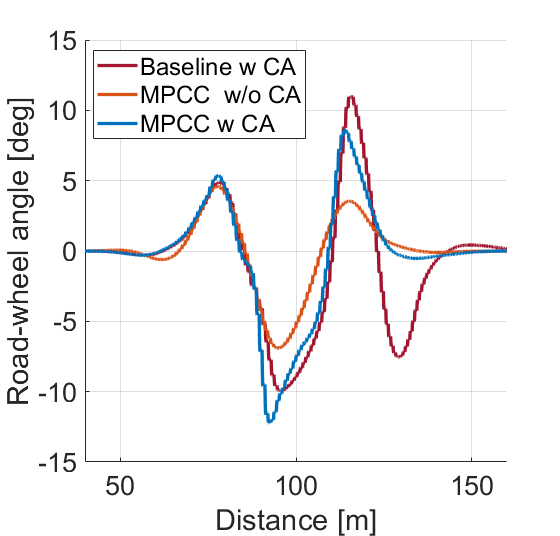}} 
    \subfloat[Longitudinal force input.\label{fig:DLFx}]{%
       \includegraphics[width=0.25\textwidth, keepaspectratio]{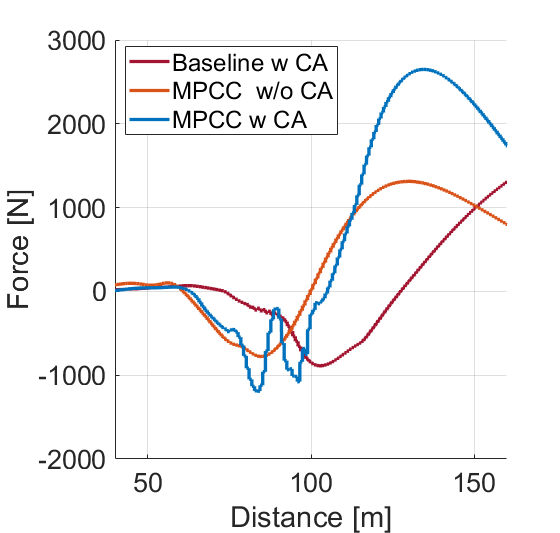}}
    \subfloat[G-G diagram.\label{fig:DLGG}]{%
       \includegraphics[width=0.25\textwidth, keepaspectratio]{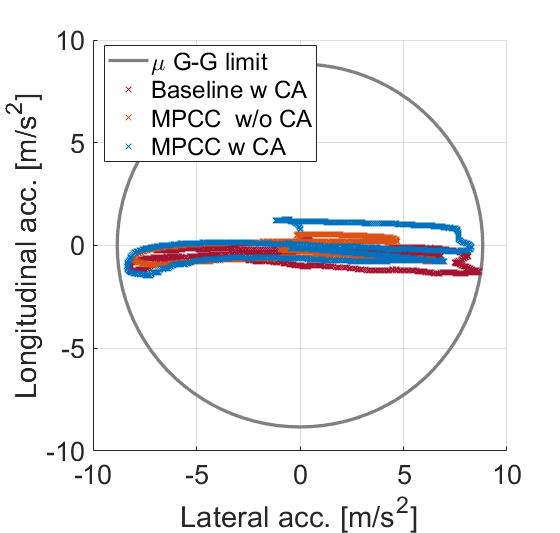}} \hfill
    \caption{States and control inputs for the double lane change manoeuvre.}
    \label{fig:DLAnalysis}
\end{figure}

Fig. \ref{fig:DLTraje} shows the vehicle trajectories obtained by three different controllers, the proposed MPCC with and without collision avoidance (CA) objective and the baseline controller for CA based on an MPC \cite{brown2019coordinating}. The MPCC without CA causes the vehicle to collide with the first obstacle of the double lane change, as visible from the negative V2O distance, see Fig. \ref{fig:DLRO}. The collision happens because the controller prioritises vehicle stability and path tracking over obstacle avoidance, even when the desired trajectory passes dangerously close to an obstacle. This situation can happen when the path tracker and the vehicle stability controller affect the tracking performance without taking into account the position of obstacles. Alternatively, when the desired trajectory is unfeasible due to a mismatch in the handling limit of the motion planner and path-tracking prediction model. Vice versa, the MPCC and the baseline, both with CA prioritisation, successfully avoid both obstacles, see Fig. \ref{fig:DLRO}. Despite the successful avoidance, the replanning around the first obstacle generated by the baseline negatively affects the rest of the manoeuvre. Thus, the vehicle trajectory has two noticeable overshoots at \SI{125}{m} and \SI{135}{m}, which causes the vehicle to enter inside the unsafe area near the second obstacle and to the right road edge, respectively Fig. \ref{fig:DLRO} and Fig. \ref{fig:DLRE}. The MPCC with CA can avoid both obstacles and keep the vehicle outside the unsafe area. This shows that CA prioritisation comes at the cost of increasing the path and velocity tracking errors and decreasing vehicle stability, pushing the vehicle to the boundaries of the handling limit. 
For this reason, the baseline creates a sideslip angle peak of \SI{9}{deg} at \SI{130}{m}, see Fig. \ref{fig:DLBeta}. The baseline needs to counter-steer to return the vehicle to linear behaviour; see Fig. \ref{fig:DLDelta}. On the contrary, the MPCC with CA has a higher stability margin than the baseline by reducing the vehicle sideslip angle peaks and allowing the vehicle to drive through the manoeuvres at a higher velocity, see Fig. \ref{fig:DLVel}. Considering Fig. \ref{fig:DLFx}, the MPCC with CA begins braking \SI{20}{m} ahead of the baseline MPC, so the vehicle velocity at the entry of the corner is lower, and the vehicle can release the brake to fully exploit the lateral tyre force. However, the baseline applies hard braking during the central part of the corner, reducing the tyre's capability of generating lateral tyre force. Thus, the vehicle exceeds its handling limit obliging the controller to brake for a longer time and to stabilise the vehicle. \compressParag\newline\indent
The MPCC applies braking before the baseline for two reasons: the Cartesian reference frame employed by the MPCC does not overestimate the V2O distance. The second reason is that the MPCC performs the tracking computing the contouring and the lag error. When the vehicle needs to replan the desired trajectory to avoid an obstacle, the MPCC controller tends to reduce the velocity because it is the only way to decrease the lag error and, consequently, the contouring error. However, the baseline MPC does not compute the lag error, so it decreases the vehicle velocity later when it faces a contouring error during the corner. However, it should be noted that the lag error in the MPCC brings complexity to the cost function weights' tuning due to the strong coupling between velocity and path tracking. Considering the available road friction coefficient, Fig. \ref{fig:DLGG} shows that all the controllers reach the maximum lateral acceleration. However, the controllers do not fully exploit the maximum braking capabilities.

\section{Conclusion}
This paper proposed an innovative approach for obstacle avoidance in automated vehicles driven at the handling limit. A non-linear MPCC integrates motion planner, path tracking and vehicle stability objectives into a single controller, prioritising obstacle avoidance in emergency situations. In a double-lane change manoeuvre, the MPCC successfully avoids two obstacles, shortly re-planning the target trajectory from the behaviour planner. At the same time, the same controller without CA prioritisation collides with the obstacles. The CA prioritisation comes with decreased path-tracking performance and increased vehicle sideslip angle peaks up to \SI{3}{deg}. However, the vehicle remains stable and manoeuvrable along the double-lane change. The state-of-the-art baseline also avoids the two obstacles but cannot keep the vehicle outside the unsafe area close to the two obstacles. Furthermore, it loses vehicle stability reaching a sideslip angle peak equal to \SI{9}{deg}. Future works involve the implementation of the proposed MPCC in an experimental vehicle and analysing its performance in different road conditions. 
\vspace{2.5mm}
\newline
\textbf{Acknowledgement.} The Dutch Science Foundation NWO-TTW supports the research within the EVOLVE project (nr. 18484).
%
%
%
\bibliographystyle{splncs03_unsrt}
\bibliography{references.bib}

\begin{thebibliography}{10}
\providecommand{\url}[1]{\texttt{#1}}
\providecommand{\urlprefix}{URL }

\bibitem{brown2019coordinating}
Brown, M., Gerdes, J.C.: Coordinating tire forces to avoid obstacles using
  nonlinear model predictive control. IEEE Transactions on Intelligent Vehicles
   5 (2019)

\bibitem{Falcone2008Hiera}
Falcone, P., Borrelli, F., Tseng, H.E., Asgari, J., Hrovat, D.: A hierarchical
  model predictive control framework for autonomous ground vehicles. American
  Control Conference  (2008)

\bibitem{funke2016collision}
Funke, J., Brown, M., Erlien, S.M., Gerdes, J.C.: Collision avoidance and
  stabilization for autonomous vehicles in emergency scenarios. IEEE
  Transactions on Control Systems Technology  25 (2016)

\bibitem{lenssen2023combined}
Lenssen, D., Bertipaglia, A., Santafe, F., Shyrokau, B.: Combined path
  following and vehicle stability control using model predictive control. Tech.
  rep., SAE Technical Paper (2023)

\bibitem{Gao2014Robust}
Gao, Y., Gray, A., Carvalho, A., Tseng, E., Borrelli, F.: Robust nonlinear
  predictive control for semiautonomous ground vehicles. American Control
  Conference  (2014)

\bibitem{chowdhri2021integrated}
Chowdhri, N., Ferranti, L., Iribarren, F., Shyrokau, B.: Integrated nonlinear
  model predictive control for automated driving. Control Engineering Practice
  106 (2021)

\bibitem{liniger2015optimization}
Liniger, A., Domahidi, A., Morari, M.: Optimization-based autonomous racing of
  1: 43 scale rc cars. Optimal Control Applications and Methods  36 (2015)

\bibitem{brito2019model}
Brito, B., Floor, B., Ferranti, L., Alonso-Mora, J.: Model predictive
  contouring control for collision avoidance in unstructured dynamic
  environments. IEEE Robotics and Automation Letters  4 (2019)

\bibitem{Bertipaglia2022Two}
Bertipaglia, A., Shyrokau, B., Alirezaei, M., Happee, R.: A two-stage bayesian
  optimisation for automatic tuning of an unscented kalman filter for vehicle
  sideslip angle estimation. In: IEEE Intelligent Vehicles Symposium (2022)

\bibitem{FORCESPro}
Domahidi, A., Jerez, J.: Forces professional. Embotech AG,
  url=https://embotech.com/FORCES-Pro (2014--2019)

\end{thebibliography}
\end{document}